\documentclass[sigconf,nonacm]{acmart}%anonymous

\usepackage{booktabs} % For formal tables

\usepackage{multirow}
\usepackage{afterpage}

\usepackage[utf8]{inputenc}
\usepackage[english]{babel}

\usepackage{hyperref}
\usepackage{url}
\usepackage{soul}
\usepackage{color}
\usepackage{multirow}
\usepackage{graphicx}
\usepackage{longtable}
\usepackage{algorithm}
\usepackage{enumitem}
\usepackage[noend]{algpseudocode}

\copyrightyear{2021} 
\acmYear{2021} 
\setcopyright{acmcopyright}\acmConference[GECCO '21 Companion]{2021 Genetic and Evolutionary Computation Conference Companion}{July 10--14, 2021}{Lille, France}
\acmBooktitle{2021 Genetic and Evolutionary Computation Conference Companion (GECCO '21 Companion), July 10--14, 2021, Lille, France}
\acmPrice{15.00}
\acmDOI{10.1145/3449726.3463138}
\acmISBN{978-1-4503-8351-6/21/07}

\begin{document}

\title{Graph-Aware Evolutionary Algorithms for Influence Maximization}
% \titlenote{}
% \subtitle{}
% \subtitlenote{}

\author{Kateryna Konotopska}
\affiliation{%
%Department of Information Engineering and Computer Science, 
  \institution{University of Trento}
  \city{Trento}%Povo
  \country{Italy}
  %\streetaddress{Via Sommarive 9}
  %\postcode{38123}
}
\orcid{}
\email{konotopska.k@gmail.com}

\author{Giovanni Iacca}
\affiliation{%
%Department of Information Engineering and Computer Science, 
  \institution{University of Trento}
  \city{Trento}%Povo
  \country{Italy}
  %\streetaddress{Via Sommarive 9}
  %\postcode{38123}
}
\orcid{0000-0001-9723-1830}
\email{giovanni.iacca@unitn.it}

%\shortauthors{}

\newcommand{\NUMRUNS}{10}
%TODO: Actually for most configurations we have 5 runs only.
%TODO: A simple visual comparison of performance without any statistical analysis of them cannot be used to build sound conclusions.
%TODO: I would not investigate the influence of single modification of the algorithm, as the combination (surely not linear) of the different effects could result beneficial and/or interesting. Thus, a series of t-tests (or Wilcoxon tests) would probably not be adequate to analyze the results. -> Uncomment "Best results combination" if space allows that.
%TODO: In the conclusions, the authors declare that "the most important progress was achieved by limiting the search space of the EA by means of node filtering" but in the "experiments" section there is space only for a few lines for this method and the obtained results are not adequately compared than the other methods.

\begin{abstract}
Social networks represent nowadays in many contexts the main source of information transmission and the way opinions and actions are influenced. For instance, generic advertisements are way less powerful than suggestions from our contacts. However, this process hugely depends on the influence of people who disseminate these suggestions. Therefore modern marketing often involves paying some targeted users, or \emph{influencers}, for advertising products or ideas. Finding the set of nodes in a social network that lead to the highest information spread --the so-called Influence Maximization (IM) problem-- is therefore a pressing question and as such it has recently attracted a great research interest. In particular, several approaches based on Evolutionary Algorithms (EAs) have been proposed, although they are known to scale poorly with the graph size. In this paper, we tackle this limitation in two ways. Firstly, we use approximate fitness functions to speed up the EA. Secondly, we include into the EA various graph-aware mechanisms, such as smart initialization, custom mutations and node filtering, to facilitate the EA convergence. Our experiments show that the proposed modifications allow to obtain a relevant runtime gain and also improve, in some cases, the spread results.
\end{abstract}





\maketitle

%----------------------------------------------------

\section{Introduction}
\label{sec:intro}

%This is possibly the key difference between searching on a search engine versus searching on a social network: other people's opinions --rather than priorities calculated by a search algorithm-- influence our choices.
Online social networks have become a relevant part in the daily life of the majority of people. Social networks are used every day to maintain contact with other people, read news, search for a job, watch movies, or do shopping. All these things are usually done based on what other people say about products, jobs, events, and so on. More formally, when the activity of a user $u$ implies an action of the user $v$, there is an edge $u$ $\rightarrow$ $v$ in the graph representing the social network. This edge might indicate, for example, a voter supporting a candidate for an election, or the fact that if a customer $u$ buys a product, the same product will be bought by the user $v$. In general, the notation $u$ $\rightarrow$ $v$ indicates that information spreads from $u$ to $v$ with a certain probability. In case of success, $v$ is said to be \emph{activated} by $u$. As we will see later, different probabilistic models, called \emph{diffusion models}, are now well-established tools to study how influence propagates over social networks.

%NP-complete
When one wants to find which set of nodes has more chances of producing the maximum possible number of activations in a social network, we are dealing with a problem known as the \emph{Influence Maximization} (IM) problem. This is a crucial problem for instance to win political campaigns, perform targeted marketing to trigger word-of-mouth, etc. Usually, one has a limited budget on the number of nodes that can be initially activated (the so-called \emph{seed} nodes). This is a particularly challenging combinatorial problem that, being proved to be NP-hard \cite{Kempe2003Maximizing}, has attracted a great research interest in the past few years. As we will briefly summarize in Section \ref{sec:related_works}, various algorithms have been recently proposed for solving the IM problem, either heuristics with provable guarantees, or metaheuristics such as Evolutionary Algorithms (EAs). However, while several successful techniques exist, these usually suffer from the so-called \emph{curse of dimensionality}, i.e., they either require too much time to converge, or produce limited-quality results.

Here, we aim at improving the performance of EAs to solve the IM problem. To do that, we first investigate two different approximations of the influence spread simulation. Then, we introduce several graph-aware mechanisms to facilitate the EA convergence. These contributions represent the main novelty of this work: to the best of our knowledge, a thorough investigation of these directions has not been done in the current literature. Overall, our experiments show that the proposed modifications produce a relevant runtime gain and also improve, in some cases, the spread results.

%We show that with the proposed improvements, EAs are competitive, also in terms of compute time, against the state-of-the-art heuristics, while providing a number of different equally-good solutions.

The rest of the paper is organized as follows. In the next Section, we provide the background concepts. Section \ref{sec:related_works} gives an overview of the state-of-the-art in the field of IM. Sections \ref{sec:ea_design} describes our methods, while Section \ref{sec:experiments} presents our experimental setup and the numerical results. Finally, Section \ref{sec:conclusions} summarizes our main findings and discusses the possible future work directions.

%----------------------------------------------------

\section{Background}
\label{sec:background}

As said earlier, the IM problem consists in finding the set of nodes that will lead to the maximal number of node activations according to a given influence spread model. More specifically, the problem can be formulated as follows: given a network, represented with a graph $G$, a diffusion model $m$, and a budget $k$, find a set of $k$ initially active nodes, such that the influence spread is maximized. Here, we use the Independent Cascade (IC) and Weighted Cascade (WC) models \cite{Kempe2003Maximizing}. In IC, at each timestep $t$ a node can be either active or inactive. A node can transit from inactive to active, but it cannot become inactive again. At the beginning, only the nodes in the seed set $S$ are in the active state. Then, at each timestep each active node $u$ has a chance of activating its neighboring nodes $v$ (if the edge $u \rightarrow v$ exists) with a probability $p$, which is the same for all the nodes in the graph and is a given parameter. The simulation stops once a convergence condition is met, see Algorithm~\ref{alg:monteCarloMaxHop} in Section~\ref{sec:approximations}. Influence spread is thus given by the number of active nodes at the end of the simulation. In WC, influence spread is calculated in the same way as in IC, the only difference being the probability of node activation: given a node $v$, the probability of being activated by one of its neighboring nodes $u$ is $1/(indegree(v))$. 

%----------------------------------------------------

\section{Related works}
\label{sec:related_works}

The existing approaches for solving the IM problem can be divided into two main categories: approximation algorithms with provable guarantees, and metaheuristics. The former provide good results, but their long runtime make them impractical on large-scale networks. The latter compromise runtime with approximate solutions, although they suffer from curse of dimensionality. We provide below an overview of the related works from both categories.

%----------------------------------------------------

\subsection{Approximation algorithms}%with provable guarantees

The IM problem was formulated for the first time as a combinatorial optimization problem in \cite{Kempe2003Maximizing}, and proven to be NP-hard. The authors also proposed a greedy algorithm that yields an ($1 - 1/e - \epsilon$) approximation. The submodular property of the objective function was later used in the CELF algorithm \cite{Leskovec2007Cost}, producing a 700x speedup w.r.t. a greedy approach. This property was further explored in CELF++ \cite{Goyal2011CELF++}, with an additional 35-55\% gain in runtime.%The same work also proposed a heuristic advancement of the CELF algorithm using path enumeration as objective function \cite{Goyal2011Simpath}.

More recent algorithms, namely TIM and TIM+ \cite{Tang2014Influence}, and its improved version IMM \cite{Tang2015Influence}, use a different approach to calculate the nodes with the maximum spread, based on Reverse Reachable (RR) sets. Each RR set is obtained by taking the set of nodes reached by the spread from one node (one for each RR set) and by putting them in an hypergraph as an hyperedge. The solution is then calculated iteratively by selecting the node with the highest degree in the hypergraph and removing then that node with all its incident edges. The difference between the three algorithms consists in the ways of calculating the sufficient number of RR sets and the method used to sample nodes for their generation. All three algorithms outperform both CELF and CELF++, while IMM outperforms both TIM and TIM+. A further improvement of IMM is BCT \cite{Nguyen2017Billion}, specifically designed for cost-aware targeted viral marketing, where each node has a benefit and a cost associated to it. BCT uses this information in order to first sample nodes with higher benefit for creation of the RR sets, resulting in an algorithm significantly faster than IMM.

%----------------------------------------------------

\subsection{Metaheuristics}

The first application of EAs to the IM problem was proposed in \cite{Bucur2016Influence}, in which a basic EA without any domain knowledge outperformed the greedy algorithm \cite{Kempe2003Maximizing} in terms of runtime while obtaining comparable influence spread results. Since that, a plethora of improvements were done in this direction. In \cite{Weskida2016Evolutionary}, for instance, a GPU-parallelized EA outperformed the greedy algorithm given the same runtime, taking up to 34 times less time to archive the same result. The impact of the EA parameters on the IM problem was studied instead in \cite{Weskida2018Finiding}. Another recent study \cite{garcia2020selection} investigated the effect of two selection schemes used for solving the IM problem with EAs, and found that the $(\mu, \lambda)$ scheme performs better than $(\mu + \lambda)$ when the size of the seed set increases.

Other works have investigated the use of \emph{smart initialization} of the initial EA population. In \cite{Xavier2016Populational}, different initialization strategies were compared, with PageRank-based initialization \cite{Rodrigues2018Optimization} giving the worst results. A smart initialization approach was also adopted in \cite{Gong2016Efficient}, that unlike standard high-degree initialization can guarantee a higher diversity of the solutions. In the same work, the authors used 2-hop spread \cite{Lee2014Fast} as approximated fitness function, and a similarity-based local search. Their results were quite promising, with a 10x speedup w.r.t. CELF and comparable influence results. In another work, the same authors proposed a discrete Particle Swarm Optimization algorithm (DPSO) \cite{Gong2016Influence}, which made use of 2-hop spread approximation and a smart initialization based on degree discount. DPSO outperformed CELF++ in terms of runtime, while the influence results were similar. Smart initialization techniques were also studied in \cite{Kromer2017Guided}, where each node has a probability to be inserted into the initial population proportional to its degree, and \cite{daSilva2018Influence}, producing better results than the basic EA.

Custom genetic operators were used in \cite{Michalak2018Informed}, where an informed mutation operator was introduced, based on a neural network predicting which nodes to change in a solution using centrality metrics as inputs. In \cite{Zhang2017Maximizing}, the EA premature convergence problem was handled using multi-population competition with specific crossover and mutation operations among populations. In other studies the IM problem was formulated instead as a multi-objective problem. In \cite{Guo2019Multi} a multi-objective differential evolution was used to maximize influence while minimizing the information delivery cost. Similarly, a multi-objective EA was used in \cite{Bucur2017Multi-Objective} to maximize influence while minimizing the size of the seed set.

%----------------------------------------------------

%there may be cases in which we may produce no new solutions. By performing the crossover as described above we still could end up with two children which are symmetrical to their parents. This can happen when the parents have all nodes in common except one. In order to produce two distinct nodes in such situation

%This operation is also helpful to increase population nodes diversity when it has particularly low values (we will define this metric later).

\section{Methods}
\label{sec:ea_design}

We now explain the main modifications we applied to the basic EA proposed in \cite{Bucur2016Influence}. This algorithm uses a direct encoding in which each individual genotype is simply a vector of IDs of nodes in the social network graph, i.e., a seed set. Initialization is done by randomly selecting a seed set of $k$ nodes from the graph. At each generation, parents are selected by performing tournament selection, which then reproduce by means of one-point crossover operator. Differently from \cite{Bucur2016Influence}, the crossover operator adopted here has two additional constraints: 1) to produce solutions without node repetitions, and 2) to produce solutions different from those given in input, in order to improve diversity. To satisfy the former condition, we swap only nodes which are not in common between the two parents, while maintaining the other nodes in common. Regarding the latter property, we simply force a mutation of at least one random node in each child produced by crossover. This is then followed by a random mutation that simply resets each node, with a given probability, to one of the other possible node IDs (excluding those already present in the seed set). The new population is then obtained with generational replacement with elitism. The evolution is continued until a maximum number of generations is reached, or stopped earlier if there is not any improvement for $\alpha$ consecutive generations, where $\alpha$ was set to 10\% of the maximum number of generations. In our experiments, we set \emph{population\_size}=100, \emph{generational\_budget}=100 generations, \emph{crossover\_rate}=1.0, \emph{mutation\_rate}=0.1, \emph{tournament\_size}=5, and \emph{num\_elites}=1. 
%The complete list of EA parameters we used in our experiments can be found in Table \ref{tab:EA_experimental_setup}.
This parametrization follows the one used in \cite{Bucur2016Influence}, to obtain a fair comparison with the basic EA. A specific parameter tuning, as done in \cite{Weskida2018Finiding}, is instead out of the scope of this work.

In order to improve the basic EA, we collected the best ideas from the literature, discussed in the previous Section, and put them in practice by adapting them to the algorithm. As we will see in the next Section, some of them were more successful, while some others hardly produced any improvement. Nevertheless, we report all of our results in order to provide valuable knowledge on what are the most promising directions. In a nutshell, the objectives of our work are to: 1) reduce the runtime of the fitness function, the main bottleneck of the EA, by using fitness function approximations; 2) boost the EA convergence with smart initialization; 3) improve the search efficiency by using graph-aware mutations, also including a mechanism for dynamic selection of multiple mutations; 4) reduce the EA search space by filtering the most promising nodes before starting the search. Next, we discuss these four elements in detail.

% \begin{table}[ht!]
% \centering
% \caption{Basic EA parameter settings.}
% \label{tab:EA_experimental_setup}
% \resizebox{\columnwidth}{!}{
% \begin{tabular}{|l|c|l|}
% \hline
% \textbf{Name} & \textbf{Value} & \textbf{Description} \\ \hline
% \emph{population\_size} & 100 & number of solutions in population \\ \hline
% \emph{generational\_budget} & 100 & maximum number of generations \\ \hline
% \emph{crossover\_rate} & 1.0 & crossover probability \\ \hline
% \emph{mutation\_rate} & 0.1 & mutation probability \\ \hline
% \emph{tournament\_size} & 5 & tournament size for tournament parent selection \\ \hline
% \emph{num\_elites} & 1 & number of best solutions to save from last generation \\ \hline
% \end{tabular}
% }
% \end{table}

%NOTE (Kateryna): The number of elites is different in the basic and the subsequent setups :( I changed it to 1 at a certain point because of the dynamic population experiments (when I was starting with 2 individuals, I wanted to preserve the exploration), but I did not redo all the experiments of the basic EA setup. In that case keeping more elites was more beneficial, it would be better to redo the basic setup with new defaults as well, but I think I did not do it because the HPC was down and I had to redo a lot of experiments in the last week prior to the deadline, when it was finally up.

%----------------------------------------------------

\subsection{Fitness function approximations}\label{sec:approximations}

The influence spread computation under the IC and WC models was proven to be \#P-complete in \cite{Goyal2011CELF++}, and as such it is one of the main bottlenecks for the IM algorithms. This is usually calculated by using \textbf{Monte Carlo} (MC) sampling. According to this procedure, the influence spread is calculated by simulating the information spread in the graph, using the specified diffusion model, a given number of times (usually very large, $>10.000$). The influence spread is then approximated by the mean spread across the simulations.

As seen earlier, an alternative to the expensive MC sampling is the use of RR sets \cite{Borgs2012Maximizing}. A preliminary study of surrogate models was instead conducted in \cite{bucur2018evaluating}, but the results were contrasting. More promising solutions make use of spread function approximations, which are faster than MC sampling. This is the idea we have decided to adopt in this work. In particular, we use two different approximations. The first one is the \textbf{2-hop spread} approximation function introduced in \cite{Lee2014Fast}, which as discussed earlier was used in \cite{Gong2016Efficient} because of its efficiency. According to this approximation, the influence spread is given by:
\[
\hat{\sigma}_{S} = \sum_{s \in S} \hat{\sigma}_{\{s\}}-\left(\sum_{s \in S} \sum_{c \in C_{s} \cap S} p(s, c)\left(\sigma_{c}^{1}-p(c, s)\right)\right)-\chi, 
\]
where: $\chi=\sum_{s \in S} \sum_{c \in C_{s} \backslash S} \sum_{d \in C_{c} \cap S \backslash\{s\}} p(s, c) p(c, d)$, $\sigma_{u}^{1}$ is the one-hop influence spread of node $u$, defined as $\sigma_{u}^{1}=1+\sum_{c \in C_{u}} p(u, c)$, and $C_{u}$ denotes the set of neighbors of node $u$.

The second approximation is a variation of MC sampling, called \textbf{MC max-hop}. In this case, influence is propagated up to a maximum number of hops $max\_hop$. If e.g. $max\_hop=2$, the influence is propagated up to the neighbors of the neighbors of the seed set $S$. The pseudo-code of the algorithm is shown in Algorithm \ref{alg:monteCarloMaxHop}, where setting \emph{max\_hop} to \emph{Inf} produces the original MC sampling.
%\vspace{-0.1cm}

\begin{algorithm}[ht!]
 \caption{Monte Carlo max-hop simulation.} \label{alg:monteCarloMaxHop}
 \hspace*{\algorithmicindent} \textbf{input}: seed set $S$, seed set size $k$, graph $G$, social model \emph{model},\\
 \hspace*{\algorithmicindent} \hspace{-0.8cm}number of simulations \emph{no\_simulations},\\
 \hspace*{\algorithmicindent} \hspace{-1.05cm}maximum number of hops \emph{max\_hop}\\
 \hspace*{\algorithmicindent} \textbf{output}: spread mean value \emph{avg}, spread standard deviation \emph{std}
\begin{algorithmic}[1]
\Procedure{MonteCarloMaxHopSimulation}{} 
\State $sample \gets \emph{initialize\_array(k)}$ 
\State $i \gets $1
\For{{$i \leq $ \emph{no\_simulations}}}
 \State $A \gets $ \emph{S}
 \State $B \gets $ \emph{S}
 \State $converged \gets $ \emph{False}
 \State $j \gets max\_hop$
 \While {\textbf{not} $converged$ \textbf{and} $j \leq max\_hop$} 
 \State $ nextB \gets $ \emph{()}
 \For{$n$ in $B$}
 \State $neighbors \gets $ \emph{get\_neighbors(n, G)}
 \State $not\_activated \gets $ \emph{set\_difference(neighbors, A)}
 \For{$m$ in $not\_activated$}
 \State $spread\_prop \gets $ \emph{get\_influence(n, m, model)}
 \State $p \gets $ \emph{random\_real(0, 1)}
 \If{\emph{p $>$ spread\_prop}}
 \State \emph{add\_element\_to\_set(m, nextB)}
 \EndIf
 \EndFor
 \EndFor
 \State $B \gets nextB$
 \If{\emph{is\_empty(nextB)}}
 \State $converged \gets True$
 \EndIf
 \State $A \gets $ \emph{union(A, B)}
 \State $j \gets j + 1$
 \EndWhile
 \State $sample[i] \gets $ \emph{length(A)}
 \State $i \gets $ \emph{i + 1}
 \EndFor
 \State $avg \gets mean(sample)$
 \State $std \gets $ \emph{standard\_deviation(sample)}
\EndProcedure
\end{algorithmic}
\end{algorithm}

%We compared both these methods with the sampling, in order to decide which of them is more convenient than the other in terms of their accuracy vs velocity trade-off. The main question we asked ourselves was how the size and the connectivity of the graph impacted the results given by each of the approximation functions and in which circumstances one of them can be more appropriate than the others.

%----------------------------------------------------

\subsection{Smart initialization}

Introducing custom-built solutions into the initial population is a common practice in Evolutionary Computation \cite{iacca2011super,caraffini2013super,caraffini2013cma}. As we have seen earlier, several works \cite{Xavier2016Populational,Gong2016Efficient,Gong2016Influence,Kromer2017Guided,daSilva2018Influence} have reported a positive effect of smart initialization on the IM problem. Here we consider different initialization techniques, described below.\\

\noindent{}\textbf{Single smart solution}: A single ``smart'' solution is inserted into the initial population, that is generated by taking the first $k$ nodes obtaining the highest centrality scores. We tested the following centrality metrics:
\begin{itemize}[leftmargin=*]
 \item betweenness: measures how often the node is on the shortest path among any two nodes in the graph;
 \item closeness: corresponds to the mean length of the shortest path between the node and all the other nodes in the graph;
 \item degree: the number of out-going links of the node;
 \item eigenvector: nodes neighboring with few highly-connected nodes score higher;
 \item katz: a variant of the eigenvector metric, where the distance between two nodes is measured by considering the number of possible paths among them, instead of only the shortest path.
\end{itemize}

\noindent{}\textbf{Multiple smart solutions}: A percentage of the initial population is initialized according to one of the following strategies:
\begin{itemize}[leftmargin=*]
 \item degree random: the probability of a node to be inserted into a solution is proportional to its degree;
 \item degree random ranked: the probability of a node to be inserted into a solution is proportional to its ranking position w.r.t. the nodes' degree;
 \item community degree: the general idea is that the network might have \emph{community structure} and a good solution should contain the most influential nodes from the largest communities. Here, we use the Louvain algorithm \cite{Blondel2008Fast} to perform community detection. The solutions are then built as follows: first, we select a community, with probability proportional to its size. Then, we select the node within the selected community, with probability proportional to its degree.
 %Here, we use two community detection algorithms, namely the Louvain algorithm \cite{Blondel2008Fast} and spectral clustering \cite{ng2002spectral}.
\end{itemize}
When using the initialization with both single and multiple smart solutions, the rest of the population is initialized randomly.
%repeating $k$ times these steps

%our goal was not the ranking of all possible community detection algorithms but rather a simple attempt to see if this kind of smart initialization could be a promising direction for the IM problem.

%----------------------------------------------------

\subsection{Graph-aware mutations}

We attempt to accelerate the search process by introducing some knowledge about the node properties and graph structure into the mutation operator. In particular, we considered four different mutation schemes, described below.\\

\noindent{}\textbf{Global mutation methods}: We mutate the selected node to another node randomly chosen from the graph. Table \ref{tab:global_mutations} recapitulates the node selection techniques we use with the global mutation methods.

\noindent{}\textbf{Local mutation methods}: The idea is to perform a \emph{local search} on the graph, i.e., to apply a mutation that preserves a certain proximity to the current solution (i.e., the seed set). In this case the node to be mutated is selected randomly, while the new node is chosen according to one of the criteria listed in Table \ref{tab:local_mutations}. 

% \noindent{}\textbf{Activation mutations}: This mechanism is inspired by the RR sets \cite{Borgs2012Maximizing}. While performing MC simulations we track, up to a certain degree, which nodes activated which other nodes. More specifically, for each node in the graph we save which nodes (activators) caused its activation, limited up to 2-hop distance to avoid an excessive memory usage. Each activator has its own counter, thus we can compute the most influential activator for each node. Subsequently, we use this information in two different ways:
% \begin{enumerate}[leftmargin=*]
% \item local activation mutation: a node $u$ is replaced by its most influential activator (i.e., the node with the highest activation counter on $u$);
% \item global activation mutation: a node $u$ is replaced by a random node $v$, satisfying the following constraints: 1) $u$ and $v$ never activated each other; 2) all the other nodes in the solution have never activated $v$, and vice versa.
% \end{enumerate}

\noindent{}\textbf{Combination of multiple mutations}: We have observed that some mutations are more effective than others when used in different evolution phases and when applied to networks with different degree distributions. Here, we model the dynamic selection of which mutation to use within a pool of available mutations as a non-stationary \emph{Multi Armed Bandit} (MAB) problem. To solve the MAB problem, we adopt the Upper Confidence Bound (UCB1) algorithm \cite{auer2002finite}, which promotes actions (in our case, mutations) with larger reward uncertainty. More precisely, to select the next action, the UCB1 algorithm maximizes the following expression: $Q(a) + exploration\_weight \cdot \sqrt{(2 \log t)/N_{t}(a)}$, where $Q(a)$ is the cumulative reward of the last $n$ times the action $a$ was selected ($n$ is a sliding window size), $N_t(a)$ corresponds to the number of times $a$ was chosen, and $t$ is the counter of the total action selections. To shrink the learning phase, we adopted an exponentially decaying exploration weight $exploration\_weight(g) = g^{-3}$, where $g$ is the generation counter. Note that the sliding window is used since the problem is non-stationary, i.e., the reward of a certain mutation may change during the evolutionary process. This mechanism should then produce a fair trade-off between exploration and exploitation along the search.

\begin{table}[ht!]
\centering
\caption{Selection techniques used with global mutations.}
\label{tab:global_mutations}
\vspace{-0.1cm}
\resizebox{\columnwidth}{!}{
\begin{tabular}{|p{.25\columnwidth}|p{.75\columnwidth}|}
\hline
\multicolumn{1}{|l|}{\textbf{Mutation}} & \textbf{Description} \\ \hline
{Global random} & The node to be mutated is selected randomly. \\ \hline
{Global low\newline{}degree} & The mutation probability of a nodes is inversely proportional to its degree.\\ \hline
{Global low\newline{}spread} & The mutation probability of a node is inversely proportional to its influence spread value.\\ \hline% calculated by simulating independently for each node its spread
{Global low\newline{}additional\newline{}spread} & The mutation probability of a node is inversely proportional to the increase in influence spread it produces when added to the seed set excluding it.\\ \hline
\end{tabular}
}
\vspace{-0.3cm}
\end{table}

\begin{table}[ht!]
\centering
\caption{Selection techniques used with local mutations.}
\label{tab:local_mutations}
\vspace{-0.1cm}
\resizebox{\columnwidth}{!}{
\begin{tabular}{|p{.25\columnwidth}|p{.75\columnwidth}|}
\hline
\multicolumn{1}{|l|}{\textbf{Mutation}} & \textbf{Description} \\ \hline
{Local neighbors\newline{}random} & The new node is selected randomly among the neighbors of the node to be mutated. \\ \hline
{Local neighbors\newline{}second degree} & The new node is selected among the neighbors of the node to be mutated, with probability proportional to its degree. \\ \hline
{Local neighbors\newline{}approximated spread} & The new node is selected among the neighbors of the node to be mutated, with probability proportional to its approximated spread.
\\ \hline
%Local neighbors additional spread & The new node is selected among the neighbors of the node to be mutated, with probability proportional to the influence increase produced when inserted in its place in the seed set. \\ \hline
{Local\newline{}embeddings\newline{}random} & The new node is selected randomly among the nodes closest to the node to be mutated, according to its corresponding node2vec \cite{Grover2016Node2vec} embeddings (i.e., a continuous feature representation of the graph's nodes, trained to predict the probability of nodes being neighbors).
\\ \hline
%NOTE (Kateryna): Node2vec is an extension of the word2vec \cite{mikolov2013efficient} algorithm. It learns continuous feature representation for nodes. It builds a "corpus" of random walks in the graph and uses it as an input to train the Skip-Gram model, which task is to predict, given a node, the probability of each node in the graph to be its "neighbor" node. The hidden layer of the Skip-Gram model is then used to extract the nodes embeddings vectors.
\end{tabular}
\vspace{-0.3cm}
}
\end{table}

%----------------------------------------------------

\subsection{Node filtering}

The search space of the IM problem is combinatorial, therefore most existing algorithms scale poorly with the graph size. We investigate two ways to reduce the search space by filtering the nodes in the graph before applying the EA.\\
%and/or to make it independent on the graph size,

\noindent{}\textbf{Min degree nodes}: Nodes with low degree have a very little chance of influencing the others, so it is very unlikely that they would be part of the optimal solution. Here, we filter the $n$ nodes with degree larger than a given threshold.

\noindent{}\textbf{Best spread nodes}: When considering large-scale networks, it is very likely that the optimal seed set would be made of nodes placed far enough from each other, to avoid activating the same nodes. According to this assumption, it is highly probable that the nodes in the optimal solution are isolated among them. In other words, their activations do not overlap and the resulting spread is very close to the sum of the spreads of each of the nodes calculated separately. To test this assumption, we first compute the spread generated by each node, separately, and then filter the best $n$ nodes in terms of mean spread. This technique requires the spread measures of any two nodes to be comparable with each other, in order to determine which one is bigger. Furthermore, this technique is computationally expensive since it requires to perform MC simulations for \emph{all the nodes} in the network, before running the EA. However, deciding \emph{a priori} the required number of MC simulations needed to have a statistically significant comparison of the spread produced by all nodes (averaged across multiple MC simulations, as seen earlier) is not possible. Therefore, we use the following procedure that increases iteratively the number of MC simulations, until a needed precision is reached:
\begin{enumerate}[leftmargin=*]
 \item Initialize a node set $\mathcal{N}$ with all the nodes in the graph, and set the maximum error rate $max\_error\_rate$ to a given value.
 \item Perform MC sampling of the spread of each node in $\mathcal{N}$ with a number of MC simulations (which might be different for each node) needed to ensure a maximum error rate $max\_error\_rate$ on the mean spread. The error of the mean is estimated by calculating the confidence interval of the Student's T distribution.
 \item Set $\mathcal{N}$ to the best $n$ nodes in terms of mean spread, plus all the nodes in the graph which are not statistically comparable with the node with the lowest spread among the best $n$ nodes. Decrease $max\_error\_rate$ and go to step 2.
\end{enumerate}
This loop is repeated until all the best $n$ nodes are comparable with the others. However, the sample size required to compare any pair of nodes would be too costly, thus we stop the loop as soon as the number of best nodes falls in a given range $[l, u]$. More specifically, as soon as the algorithm finds a number of best nodes $l$, and the number of nodes incomparable with those $l$ nodes is lower than $u - l$, it stops. When the algorithm terminates, it returns the $l$ best nodes, plus all the nodes which cannot be compared with them.

%----------------------------------------------------

\begin{table}[ht!]
\centering
\caption{Real-world datasets specifications.}
\label{tab:real_datasets}
\resizebox{\columnwidth}{!}{
\begin{tabular}{|l|p{.23\columnwidth}|c|c|c|c|c|}
\hline
\multicolumn{1}{|c|}{\multirow{2}{*}{\textbf{Dataset}}} & \multicolumn{1}{c|}{\multirow{2}{*}{\textbf{Graph size}}} & \multicolumn{5}{c|}{\textbf{Node degree}} \\ \cline{3-7} 
\multicolumn{1}{|c|}{} & \multicolumn{1}{c|}{} & \multicolumn{1}{c|}{\textbf{Avg}} & \multicolumn{1}{c|}{\textbf{Std}} & \multicolumn{1}{c|}{\textbf{Min}} & \multicolumn{1}{c|}{\textbf{Max}} & \multicolumn{1}{c|}{\textbf{Median}} \\ \hline
%Epinions & {75.879 nodes\newline{}508.837 edges} & 13.41 & 52.67 & 1 & 3079 & 2 \\ \hline
\multirow{2}{*}{Wiki-vote} & {7.115 nodes\newline{}103.689 edges} & \multirow{2}{*}{14.57} & \multirow{2}{*}{42.28} & \multirow{2}{*}{0} & \multirow{2}{*}{893} & \multirow{2}{*}{2} \\ \hline
\multirow{2}{*}{Amazon} & {262.111 nodes\newline{}1.234.877 edges} & \multirow{2}{*}{4.71} & \multirow{2}{*}{0.95} & \multirow{2}{*}{0} & \multirow{2}{*}{5} & \multirow{2}{*}{5} \\ \hline
\multirow{2}{*}{CA-GrQc} & {5.242 nodes\newline{}14.496 edges} & \multirow{2}{*}{5.52} & \multirow{2}{*}{7.92} & \multirow{2}{*}{1} & \multirow{2}{*}{81} & \multirow{2}{*}{3} \\ \hline
\end{tabular}
}
\end{table}

\begin{table}[ht!]
\centering
\caption{Synthetic Barab\'{a}si-Albert datasets specifications. The $n\_edges$ parameter indicates the number of edges added from a new node to the existing nodes.}
\label{tab:fake_datasets}
\resizebox{\columnwidth}{!}{
\begin{tabular}{|l|c|c|c|c|c|}
\hline
\multicolumn{1}{|c|}{\multirow{2}{*}{\textbf{Graph size}}} & \multicolumn{5}{c|}{\textbf{Node degree}} \\ \cline{2-6} 
\multicolumn{1}{|c|}{} & \multicolumn{1}{c|}{\textbf{Avg}} & \multicolumn{1}{c|}{\textbf{Std}} & \multicolumn{1}{c|}{\textbf{Min}} & \multicolumn{1}{c|}{\textbf{Max}} & \multicolumn{1}{c|}{\textbf{Median}} \\ \hline
{1.000 nodes\newline{}$n\_edges$=1} & 1.99 & 3.51 & 1 & 75 & 1 \\ \hline
{1.000 nodes\newline{}$n\_edges$=3} & 5.98 & 7.37 & 3 & 99 & 4 \\ \hline
{1.000 nodes\newline{}$n\_edges$=5} & 9.95 & 10.57 & 2 & 114 & 7 \\ \hline
{1.000 nodes\newline{}$n\_edges$=7} & 13.90 & 12.80 & 7 & 137 & 10 \\ \hline
{1.000 nodes\newline{}$n\_edges$=9} & 17.84 & 15.65 & 9 & 148 & 12 \\ \hline
{1.000 nodes\newline{}$n\_edges$=11} & 21.76 & 19.25 & 11 & 211 & 15 \\ \hline
{10.000 nodes\newline{}$n\_edges$=1} & 1.99 & 4.13 & 1 & 243 & 1 \\ \hline
{10.000 nodes\newline{}$n\_edges$=3} & 5.99 & 8.96 & 3 & 293 & 4 \\ \hline
{10.000 nodes\newline{}$n\_edges$=5} & 9.99 & 13.45 & 2 & 365 & 7 \\ \hline
{10.000 nodes\newline{}$n\_edges$=7} & 13.99 & 17.00 & 7 & 446 & 10 \\ \hline
{10.000 nodes\newline{}$n\_edges$=9} & 17.98 & 21.25 & 9 & 470 & 12 \\ \hline
{10.000 nodes\newline{}$n\_edges$=11} & 21.98 & 25.97 & 11 & 695 & 15 \\ \hline
\end{tabular}
}
\end{table}

\section{Experiments}
\label{sec:experiments}

%Each experiment used was executed in single thread due to hardware limitations.
For the experiments, we used three real-world graphs taken from the SNAP repository \cite{snapnets}, namely Wiki-Vote, Amazon and CA-GrQc. In Wiki-Vote the edges represent who-voted-whom information in elections for promoting adminship. The Amazon dataset contains co-purchased products relations. The smallest dataset, CA-GrQc, is a network of collaboration in the General Relativity and Quantum Cosmology fields. Table \ref{tab:real_datasets} contains a detailed description of each of the aforementioned graphs. For testing the function approximations, we also used 6 synthetic datasets generated by the Barab\'{a}si-Albert model \cite{Barabasi1999Emergence}, detailed in Table \ref{tab:fake_datasets}. These were created using the \texttt{networkx} library \cite{networkx} with a $random\_seed$ parameter set to $0$. Except for the function approximations experiments, where we set $p=0.1$ in the IC model and $max\_hop=2$, in all the other experiments we set $p=0.01$ and $max\_hop=3$. Each of the four proposed modifications presented above (fitness function approximations, smart initialization, graph-aware mutations, and node filtering) was tested separately. Every experimental condition was repeated for \NUMRUNS~independent runs (in the next figures, we provide mean values $\pm$ std. dev.)\footnote{Our code is available at:
\ifdefined\ANONYMOUS
\emph{link omitted for double-blind review}.
\else
\url{https://github.com/katerynak/Influence-Maximization}.
\fi}.
\subsection{Fitness function approximations}

The first part of the experimentation was aimed at verifying which fitness function approximation works best in terms of runtime and quality of results. To do that, we compared the runtime and spread (mean and std. dev. across multiple simulations) computed by MC sampling and the two approximations, i.e., 2-hop spread and MC max-hop ($max\_hop=2$), on the different datasets above, under both the IC and WC models. For each dataset and model, the three methods were executed on $100$ randomly generated seed sets, the same for all the methods. The Pearson correlation on the spread values calculated by the proposed approximations and MC sampling was then calculated.

\begin{figure*}[ht!]
\centering
\includegraphics[width=1\textwidth,trim=0 12cm 0 2cm, clip]{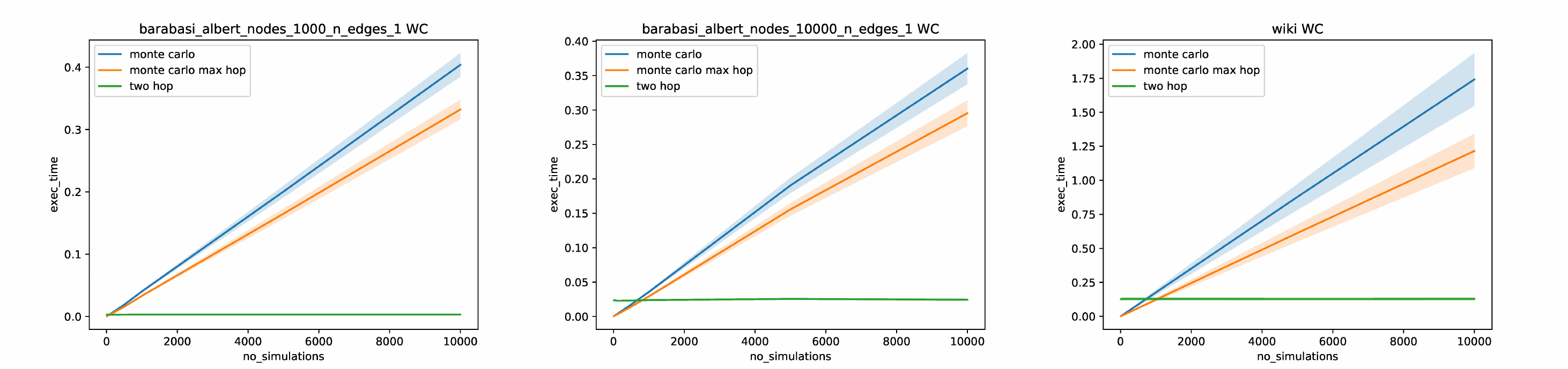}
\includegraphics[width=1\textwidth,trim=0 0 0 12cm, clip]{figures/fitness_comparisons_no_simulations.pdf}
\vspace{-1.5cm}
\caption{Spread function runtime as a function of the number of MC simulations.}
\label{fig:spread_functions_no_sim}
\end{figure*}

\begin{figure*}[ht!]
\centering
\includegraphics[width=.95\textwidth]{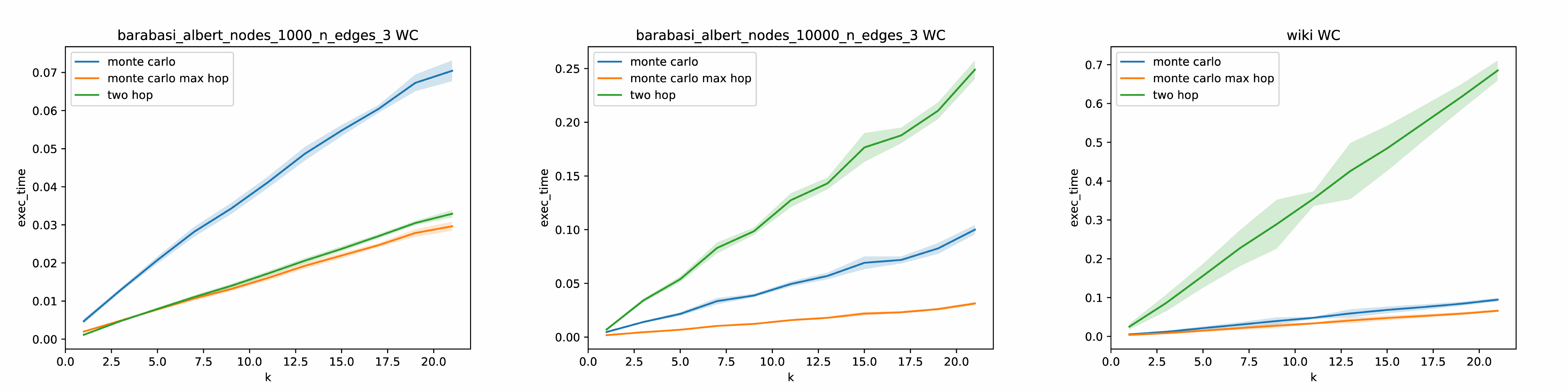}
\vspace{-0.3cm}
\caption{Spread function runtime as a function of the seed set size $k$.}
\label{fig:spread_functions_k}
\end{figure*}

Generally, we observed that both approximations in some cases had an important gain in runtime and high spread correlations w.r.t. MC sampling. The correlation of the spread values is high for both approximations ($>0.9$) for the WC model, and for the IC model with low $p$ values ($p \leq 0.1$). On the other hand, we observed a dependency of the runtime on the size and connectivity of the graph, the size of the seed set $k$, and, of course, the number of simulations used for the MC methods. Figure \ref{fig:spread_functions_no_sim} shows the runtime of the three methods on six selected datasets as the number of MC simulations varies (WC model, seed set size $k=5$). Here we can observe the linear growth of the runtime of the MC methods w.r.t the number of simulations. We can also notice that in some cases the 2-hop spread is slower not only when the number of MC simulations is low (somehow an obvious result), but also when the dataset size grows. Another thing which can be noticed is that MC max-hop is particularly useful as the dataset connectivity increases. See the Barab\'{a}si-Albert graphs in the figure, where MC max-hop obtains a lower performance gain in the top row (lower connectivity) than in the bottom row (higher connectivity). We observed this trend in all our experiments: more connected graphs yield a longer influence spread process since usually more nodes get influenced, so a truncation of this process up to some level (as in MC max-hop) can lead to larger runtime gains.%\footnote{The small fluctuation on the 2-hop curve in the Amazon WC plot is due to a different HPC configuration.}

% NOTE (Kateryna): in the Amazon WC case, the 2-hop curve changes with no_simulations since I used HPC with different nodes for calculating the exec times with different number of simulations, it may be due to that :/ 

We also observed that the runtime has, obviously, a linear relationship with the seed set size $k$. Figure \ref{fig:spread_functions_k} reports the runtime as a function of $k$ (WC model, number of MC simulations set to $100$). Here we can observe that the 2-hop spread may be convenient only if the graph size is kept small, while for the real-world cases it might be the slowest spread evaluation method.

%----------------------------------------------------

\begin{figure*}[ht!]
\centering
\includegraphics[width=1\textwidth, trim=0 0 0 10.5cm, clip]{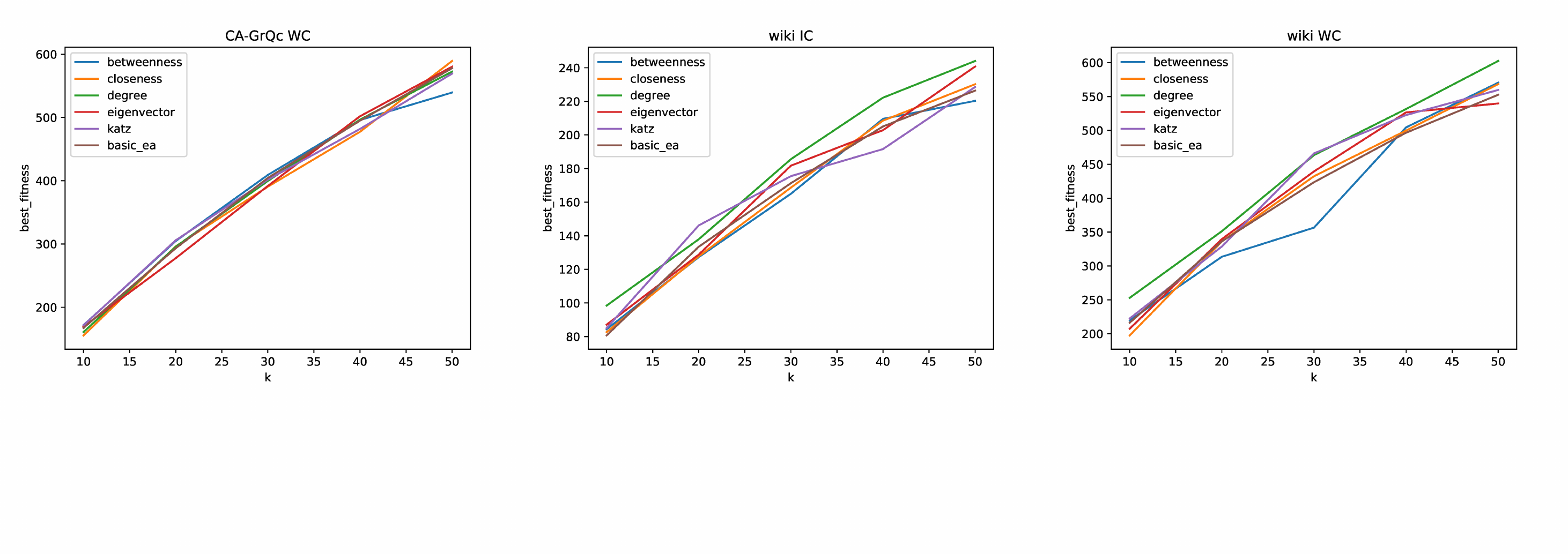}
\vspace{-2.4cm}
\caption{Single smart solution initialization techniques compared.}
\label{fig:smart_init_single}
\end{figure*}

\subsection{Smart initialization}

We tested smart initialization only on the three real-world datasets. Overall, the results indicate that smart initialization is most useful in the case of datasets that contain single, highly connected nodes.\\

\noindent{}\textbf{Single smart solution}: We were not able to run complete experiments for the Amazon dataset, given their cost in memory and/or runtime (we did not consider centrality metrics with computation time $>$ 12 hours). As shown in Figure \ref{fig:smart_init_single}, the \emph{degree} metric is a clear winner for the Wiki-Vote dataset, which is reasonable considering its degree distribution. CA-GrQc experiments did not present a clear difference between various metrics under the IC model, while when considering the WC model the \emph{betweenness} metric obtained slightly better results w.r.t. the other metrics. The only metric we could compute for the Amazon dataset was the \emph{degree} metric, but this did not bring any improvement to the basic EA (results not shown for brevity).\\

\noindent{}\textbf{Multiple smart solutions}: Due to computing resource limitations, we could not run these experiments for different values of $k$. We used instead a fixed seed set size, $k=10$, and a percentage of smart solutions of $50\%$. From Table \ref{tab:multiple_smart_solutions_comparison}, it can be seen that there is not a clear winner. On the other hand the \emph{degree random} strategies are to be preferred because of the much lower computation time requirement w.r.t. the \emph{community degree} strategy.

% \begin{table}[ht!]
% \centering
% \caption{Multiple smart solutions initialization techniques compared (for $k=10$).}
% \label{tab:multiple_smart_solutions_comparison}
% \resizebox{\columnwidth}{!}{
% \begin{tabular}{|l|c|c|c|c|c|}
% \hline
% \multicolumn{1}{|c|}{\multirow{2}{*}{\textbf{Dataset}}} & \multirow{2}{*}{\textbf{Model}} & \multicolumn{4}{c|}{\textbf{Best fitness (averaged across \NUMRUNS~runs)}} \\ \cline{3-6}
% \multicolumn{1}{|c|}{} & & \textbf{Degree random} & \textbf{Community degree} & \textbf{Degree random ranked} & \textbf{No smart. init.} \\ \hline
% {Wiki-Vote} & WC & 226.24 & 223.35 & \textbf{230.65} & 220.75 \\ \hline
% {Wiki-Vote} & IC & 93.7 & 93.21 & \textbf{94.04} & 87.92 \\ \hline
% {Ca-GrQc} & WC & \textbf{147.38} & 147.36 & 146.91 & 146.13 \\ \hline
% {Ca-GrQc} & IC & 18.64 & 18.59 & 18.59 & \textbf{18.67} \\ \hline
% {Amazon} & WC & 66.55 & 65.99 & \textbf{67.97} & 67.87 \\ \hline
% {Amazon} & IC & 10.79 & \textbf{10.83} & 10.80 & 10.82 \\ \hline
% \end{tabular}
% }
% \end{table}

\begin{table}[ht!]
\centering
\caption{Multiple smart solutions initialization techniques compared (for $k=10$).}
\label{tab:multiple_smart_solutions_comparison}
\resizebox{\columnwidth}{!}{
\begin{tabular}{|c|c|p{0.13\columnwidth}|p{0.13\columnwidth}|p{0.13\columnwidth}|p{0.13\columnwidth}|}
\hline
\multicolumn{1}{|c|}{\multirow{4}{*}{\textbf{Dataset}}} & \multirow{4}{*}{\textbf{Model}} & \multicolumn{4}{c|}{\textbf{Best fitness (averaged across \NUMRUNS~runs)}} \\ \cline{3-6}
\multicolumn{1}{|c|}{} & & \textbf{Degree\newline{}random} & \textbf{Comm.\newline{}degree} & \textbf{Degree\newline{}random\newline{}ranked} & \textbf{No\newline{}smart.\newline{}init.} \\ \hline
{Wiki-Vote} & WC & 226.24 & 223.35 & \textbf{230.65} & 220.75 \\ \hline
{Wiki-Vote} & IC & 93.7 & 93.21 & \textbf{94.04} & 87.92 \\ \hline
{Ca-GrQc} & WC & \textbf{147.38} & 147.36 & 146.91 & 146.13 \\ \hline
{Ca-GrQc} & IC & 18.64 & 18.59 & 18.59 & \textbf{18.67} \\ \hline
{Amazon} & WC & 66.55 & 65.99 & \textbf{67.97} & 67.87 \\ \hline
{Amazon} & IC & 10.79 & \textbf{10.83} & 10.80 & 10.82 \\ \hline
\end{tabular}
}
\end{table}

%----------------------------------------------------

\subsection{Graph-aware mutations}

Also in this case we tested only the Wiki-vote, CA-GrQc and Amazon graphs. We omit for brevity the detailed results, but we report below our main findings.\\

%The results of our experiments are shown in Figure \ref{fig:globalMutations}. For the Wiki-Vote dataset the execution of methods involving spread calculation were not completed because of the high computation time requirement.
\noindent{}\textbf{Global mutations}: The \emph{global low degree} mutation obtained slightly better results on the graphs presenting a low number of large degree nodes, such as Wiki-Vote and CA-GrQc, while for the Amazon dataset it led to worse results. The runtime of these methods were in general comparable, except the \emph{global low additional spread} method whose runtime grows linearly with the seed set size.

\begin{figure*}[ht!]
\centering
\includegraphics[width=.32\textwidth]{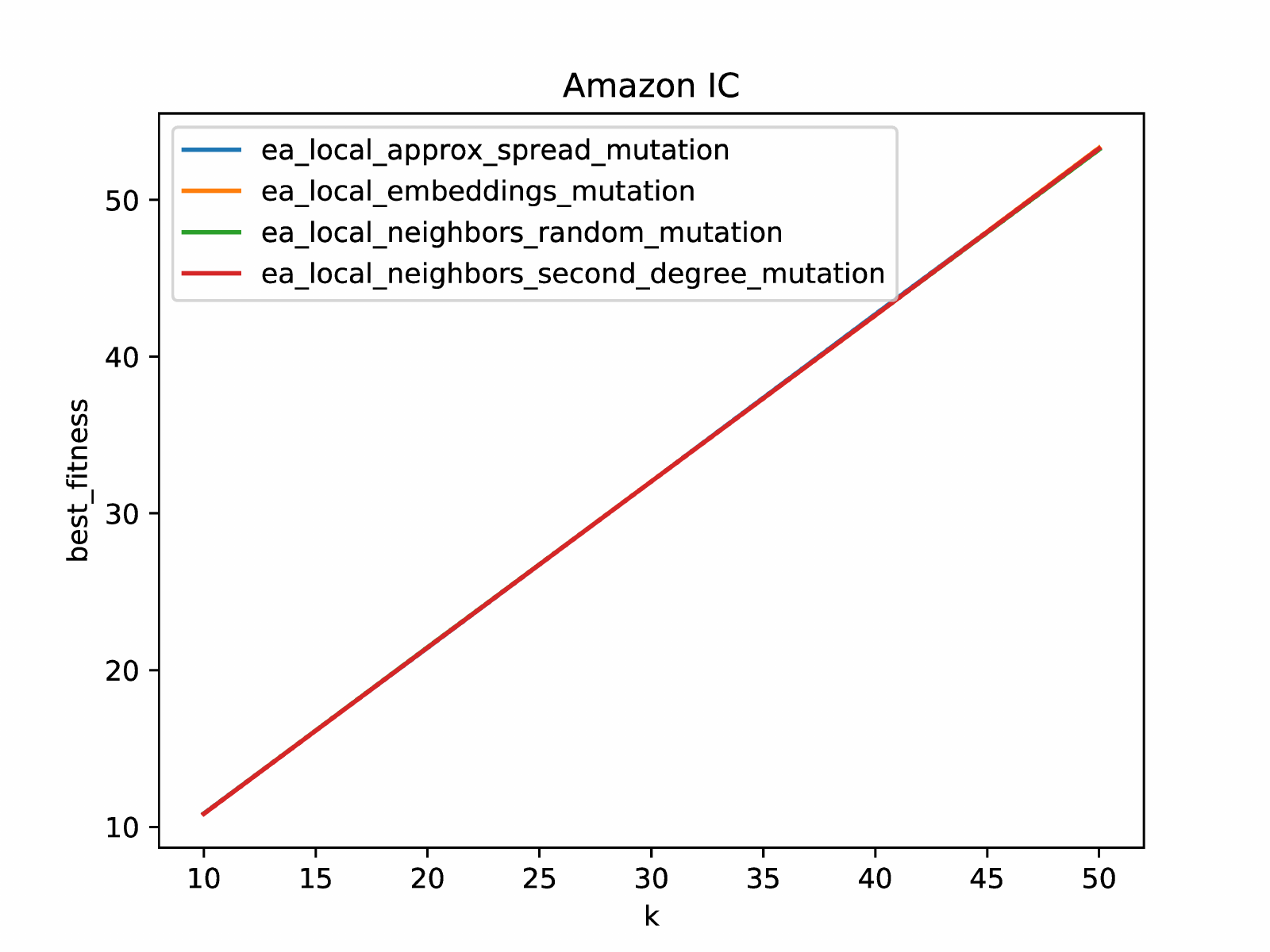}
\includegraphics[width=.32\textwidth]{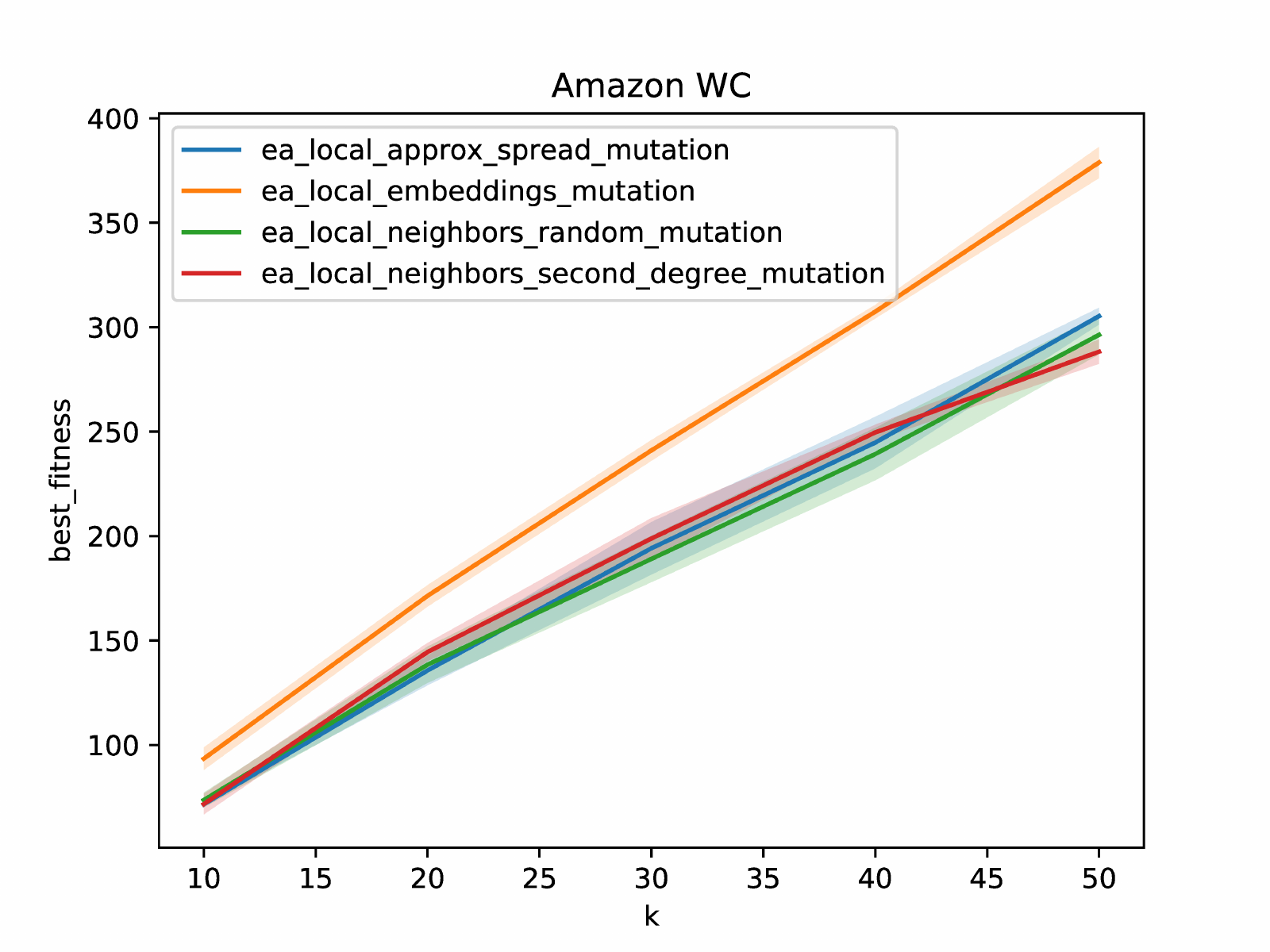}
\includegraphics[width=.32\textwidth]{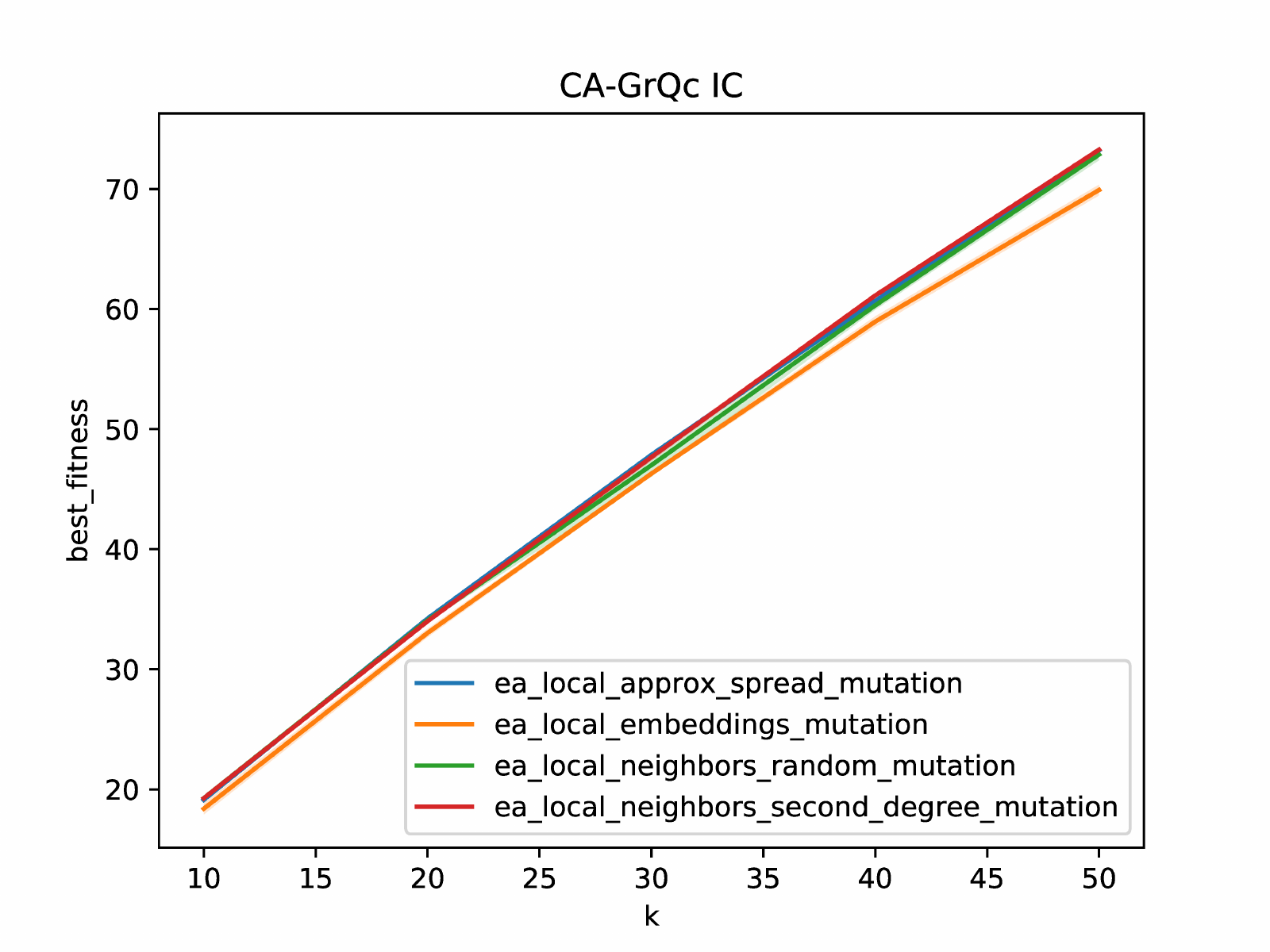}

\includegraphics[width=.32\textwidth]{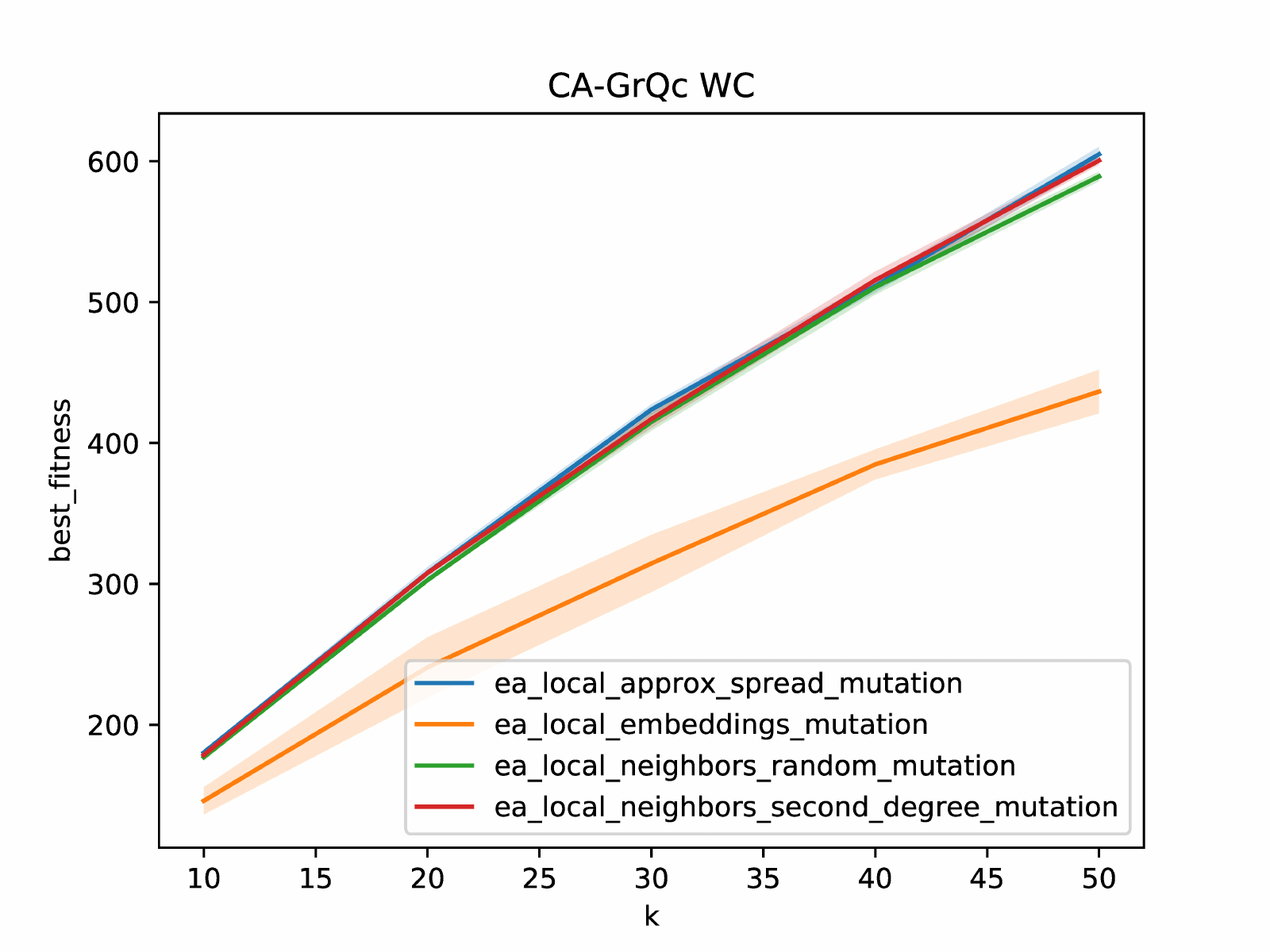}
\includegraphics[width=.32\textwidth]{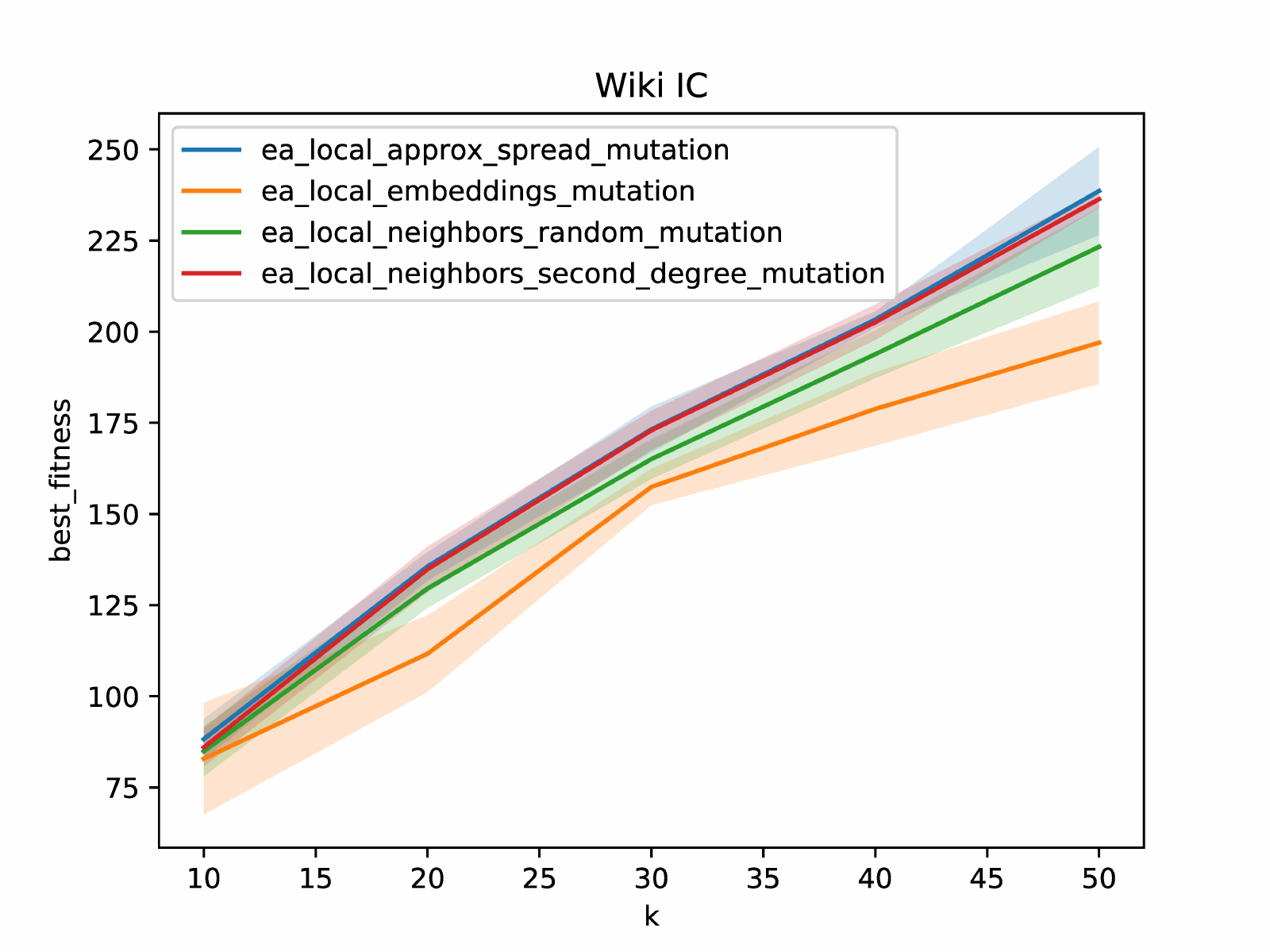}
\includegraphics[width=.32\textwidth]{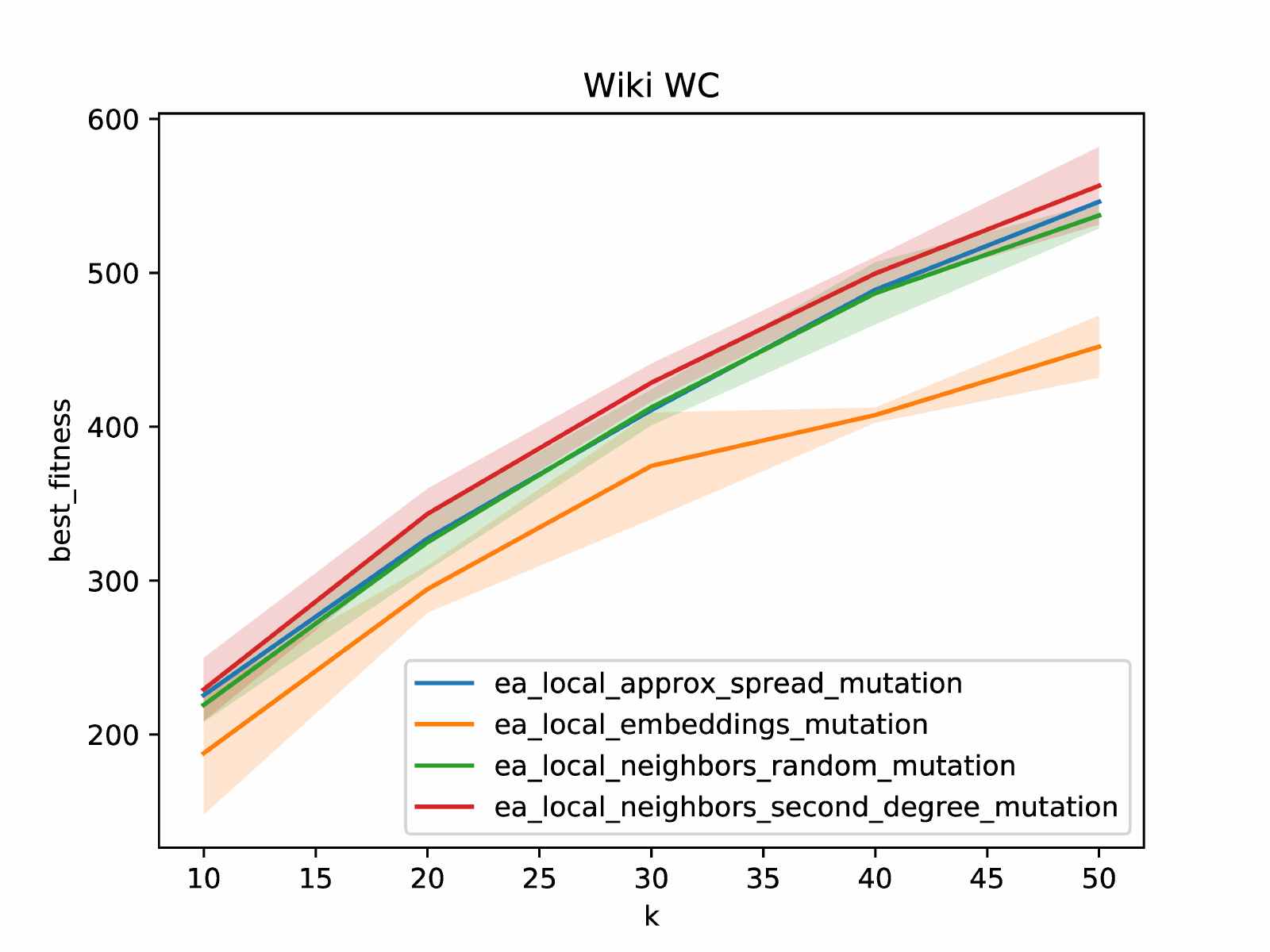}
\vspace{-0.2cm}
\caption{Local mutations compared.}
\label{fig:local_mutations}
\end{figure*}

%Before reviewing the results we should add an important detail regarding the \emph{local embeddings random} mutation: we performed some hyper-parameters search for the Amazon node2vec embeddings training, where we tried a range of different parameters for node2vec training and selected the parameters which produced embeddings which better captured the shortest path length between two nodes and nodes' degrees, while for the other datasets we just picked random parameters without any hyper-parameter search.
%Due to computing resource limitations, we could not compute the \emph{local neighbors additional spread} mutation, which, like its global variant, is particularly slow on the graphs with high-degree nodes. 
%, where we also show the results of the \emph{global random} mutation used in the basic EA. 
\noindent{}\textbf{Local mutations}: We report the results of the local mutations in Figure \ref{fig:local_mutations}. For this specific experiment, we trained node2vec on the Amazon dataset and fine-tuned its parameters. As expected then, the \emph{local embeddings random} mutation performed better on the Amazon dataset, while its results for CA-GrQc and Wiki-Vote were way worse than the other mutations: this shows that local embeddings may work well, but at a cost of an expensive and time-consuming tuning process. On the other hand, the \emph{local neighbors second degree} mutation was at least as good as the \emph{global random} mutation for all the datasets, while the \emph{local neighbors approximated spread} and the \emph{local neighbors random} mutation worked well on CA-GrQc and Wiki-Vote, but not on the Amazon dataset.

%, i.e., only the last 100 mutations were taken into account to compute the expected reward
\noindent{}\textbf{Combination of multiple mutations}: We tested the UCB1 algorithm with all the mutations in Tables \ref{tab:global_mutations}-\ref{tab:local_mutations} and a sliding window size of $100$. %As we can notice from Figure \ref{fig:multiple_mutations} for three selected cases, 
We noted that while the improvement w.r.t. the basic EA was not very significant, it was however positive for all graphs, which confirms that the dynamic selection of mutations adapts well to different cases (results not shown for brevity).
%NOTE (Giovanni): actually these experiments also involved activation mutation e local neighbors additional spread (which are now commented in the text, due to space limitations)

% \begin{figure*}[ht!]
% \centering
% \includegraphics[width=1\textwidth]{figures/globalMutations-global_comp-one_page.pdf}
% \vspace{-2cm}
% \caption{Global mutations compared.}
% \label{fig:globalMutations}
% \end{figure*}

% \begin{figure*}[ht!]
% \centering
% \includegraphics[width=.95\textwidth, trim=0 0 0 11cm, clip]{figures/multiple-one_page.pdf}
% \vspace{-2cm}
% \caption{Multiple mutations vs basic EA.}
% \label{fig:multiple_mutations}
% \end{figure*}

%----------------------------------------------------

\subsection{Node filtering}

Filtering was tested only on the real-world graphs. Below we report our findings.\\

\noindent{}\textbf{Min degree nodes}: Min-degree values of 1, 2, 3 and 4 were tested. We obtained a visible improvement w.r.t. the basic EA only on the Wiki-Vote dataset, while in the other cases the high std. dev. did not permit to define a clear winner.

\begin{figure*}[ht!]
\centering
\includegraphics[width=0.95\textwidth,trim=0 0 0 3.5cm, clip]{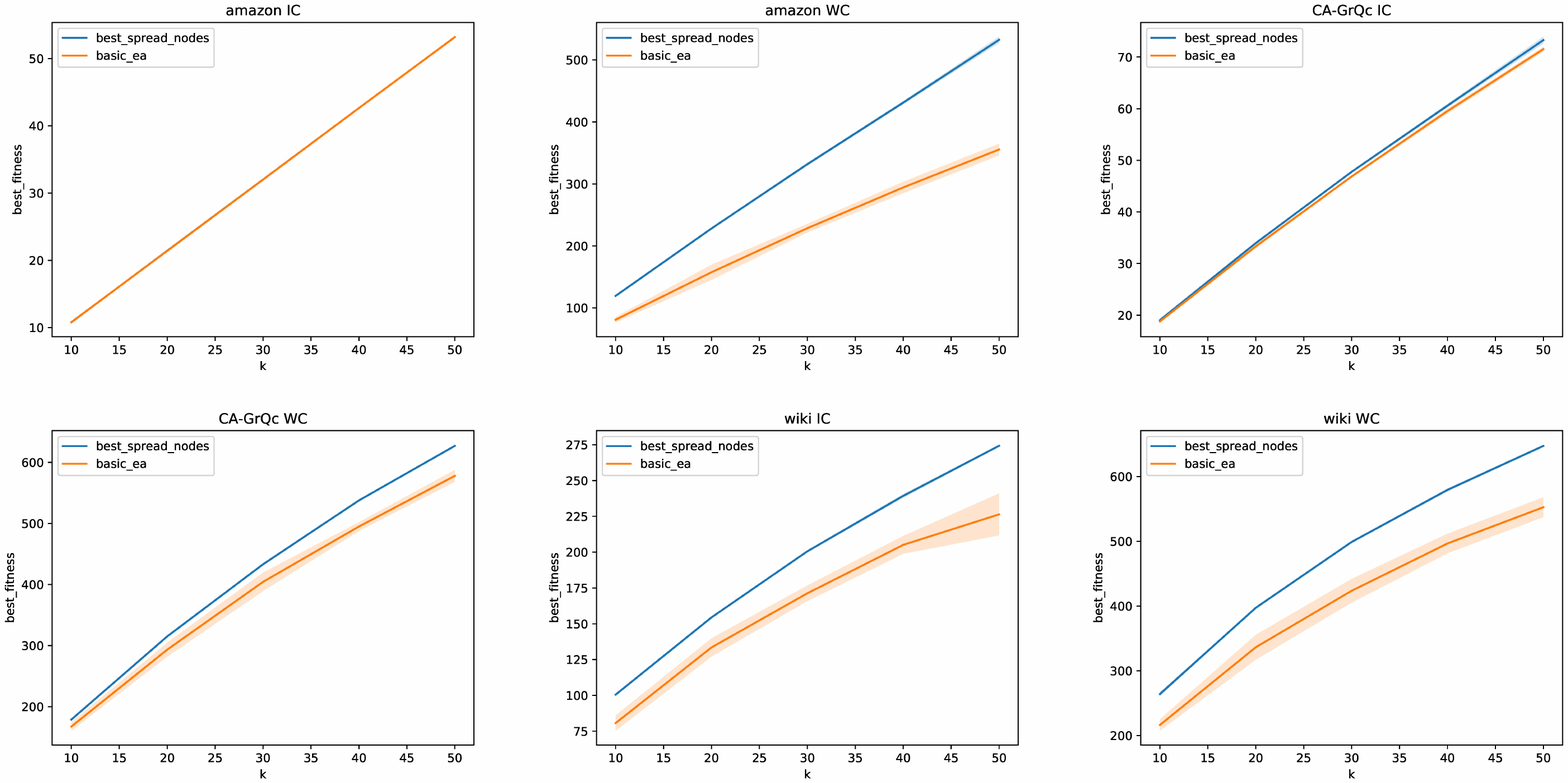}
\vspace{-1.9cm}
\caption{Best spread node filtering compared.}
\label{fig:best_spread_nodes_fig}
\end{figure*}

\noindent{}\textbf{Best spread nodes}: We used MC max-hop with $max\_hop=2$ to approximate the spread of each single node; $max\_error\_rate$ was set initially to $0.8$ and subsequently decreased of $0.1$ at each iteration. We set $l=10^9$ and $u=10^{11}$, in order to have the same search space size regardless of $k$. From Figure \ref{fig:best_spread_nodes_fig}, we can see that this method significantly improved the basic EA (up to $40\%$).

%----------------------------------------------------

\section{Conclusions}
\label{sec:conclusions}

We improved a basic EA applied in previous research to the IM problem with various graph-aware enhancements aimed at reducing the algorithm runtime. Using MC max-hop instead of MC sampling to evaluate the influence spread permitted a significant computational time saving, and allowed the possibility to conduct a variety of experiments. The most important progress was achieved by limiting the search space of the EA by means of node filtering, a method which selected the most promising nodes in terms of information spread before running the evolutionary search: the limited number of node combinations allowed in fact the EA to better scale with the increasing graph size. Moreover, the combination of several local and global graph-aware mutations permitted the EA to adapt to graphs with different structure or connectivity. 
%Dynamic population size was useful to reduce fitness evaluation costs and smart mutations computations. We compared our results with the CELF algorithm and obtained a significant runtime gain with comparable influence spread results for the Amazon dataset, and improved spread results with longer runtime for the other datasets. The important advantage of the EA is the diversity of proposed solutions it presents with comparable fitnesses. 

There are, on the other hand, some limitations in the proposed mechanisms. For instance, MC max-hop may be not a good choice in case of networks with scarce connectivity or under the IC model with high probability values. Furthermore, node filtering is valid only under the assumption that the most influential nodes are distant, in the sense that they do not influence the same nodes. So this method may perform worse on small-scale networks where the optimal seed set may include nodes with common spread influence. Additionally, the runtime of the metrics which involve the computation of the influence spread, as well as the degree-based mutations, is proportional to the fitness of the solution. So, the runtime of the algorithm is proportional to the graph connectivity and a large number of fitter candidates found during the search may imply an increasingly high computation cost.

There is still further research to do on graph-aware mutations, which would probably work better with a small number of pre-filtered nodes. In particular, the success of the node2vec embeddings mutation (although with some hyper-parameter search) suggests that this might be a promising direction to explore more in depth, also because differently from the other tested mutations this kind of mutation does not need node-specific expensive computations.

\clearpage

\bibliographystyle{ACM-Reference-Format}
\bibliography{main_gecco} 

\end{document}